\documentclass[11pt]{article}

\usepackage[preprint]{acl}

\usepackage{times}
\usepackage{latexsym}
\usepackage[T1]{fontenc}
\usepackage[utf8]{inputenc}
\usepackage{microtype}
\usepackage{inconsolata}
\usepackage{graphicx}
\usepackage{amsmath}
\usepackage{amssymb}
\usepackage{xcolor}
\usepackage{booktabs}
\usepackage{subcaption}

\title{ToolGate: Token-Efficient Pre-Call Control for Tool-Augmented Vision-Language Agents}

\author{
Anjie Liu\textsuperscript{1}
\quad
Yan Song\textsuperscript{2,3}
\quad
Zhixun Chen\textsuperscript{1}
\quad
Ziqin Gong\textsuperscript{1}
\quad
Zhongwei Yu\textsuperscript{1}
\quad
Jun Wang\textsuperscript{2}
\\[0.5em]
\textsuperscript{1}The Hong Kong University of Science and Technology (Guangzhou)
\\
\textsuperscript{2}University College London
\\
\textsuperscript{3}AI Lab, The Yangtze River Delta
}

\begin{document}
\maketitle

\begin{abstract}
Tool-augmented vision-language agents can acquire external perceptual evidence through OCR, detection, segmentation, and other tools, but executing every proposed tool call is costly and sometimes unnecessary.
We study the pre-call control problem: after a ReAct-style VLM agent proposes a perceptual tool call, should the call be executed, or skipped before its output enters the context?
Across five benchmarks, we find that the baseline agent exhibits poor local selectivity: helpful and harmful calls occur at similar rates (11.8\% vs.\ 9.9\%), while most calls do not change the immediate forced-answer prediction.
We introduce \textsc{ToolGate}, a lightweight external controller that predicts execute/skip decisions from trajectory text and simple structural features.
Across two Qwen3-VL backbones, ToolGate reduces token cost to 64--69\% of the unrestricted ReAct baseline while preserving average accuracy in cross-domain settings.
With matched-domain trajectory training on Qwen3-VL-30B, it further improves average accuracy by 1.65 points.
These results show that tool-augmented VLM agents benefit not only from better perceptual tools, but also from explicit control over when tool outputs are worth paying for.
\end{abstract}

\section{Introduction}
\label{sec:intro}

Tool-augmented agents extend base models by allowing them to call external modules during reasoning, including search engines, calculators, APIs, and perceptual tools~\citep{yao2023react,schick2023toolformer,lu2023chameleon}.
In vision-language settings, this paradigm has been instantiated by systems that compose VLMs or language-model planners with OCR, detection, segmentation, cropping, and other visual experts~\citep{yang2023mmreact,wu2023visualchatgpt,gupta2023visualprogramming,suris2023vipergpt}.
Such tools are useful because many visual questions require evidence that is difficult to extract from a single global image pass: small text may be illegible, relevant objects may occupy tiny regions, and spatial relations may require specialized perception.

However, tool access introduces a second decision problem.
At each reasoning step, the agent must decide not only \emph{which} tool to call, but whether a proposed call is worth executing at all.
A tool call can provide decisive evidence, but it can also be redundant, consume tokens, add latency, or introduce misleading perceptual information into the context.
Thus, tool use is not only a capability problem; it is also an inference-time control problem.
Given a trajectory prefix and a proposed perceptual tool invocation, should the agent execute this specific call, or continue without adding its output to the context?
We call this ability \emph{local tool-call selectivity}.

We investigate this phenomenon in ReAct-style vision–language agents.
Across five benchmarks evaluated with Qwen3-VL-30B~\citep{bai2025qwen3vl}, the baseline agent produces $15{,}782$ labeled tool invocations.
By inserting forced-answer probes immediately before and after each invocation, we observe that only 11.8\% of calls directly shift the model’s prediction from incorrect to correct, whereas 9.9\% flip it from correct to incorrect.
\emph{The remaining 78.3\% leave the immediate forced-answer argmax unchanged after the tool output is incorporated.}
This does not mean that tools are useless.
Some calls clearly help: these are the cases where the agent needs external evidence to recover the correct answer.
The problem is that such calls are mixed with many others that either leave the answer unchanged or make it worse.
Since every executed call adds tool-output tokens, interaction tokens, and extra context for later reasoning, blindly executing all proposed calls is inefficient.
This motivates a pre-call controller that keeps useful calls but skips calls that are likely to add cost without improving the answer.

We propose \textsc{ToolGate}, a lightweight external controller that operates before each proposed tool call.
Given the current trajectory and pending call, ToolGate decides from the interaction history so far: the question, previous reasoning, previous tool calls and outputs, the pending tool name and arguments, and simple metadata such as step index and tool type.
It does not access hidden states, image features, decoding log probabilities, or tool outputs, and it does not modify the VLM, prompt, or tool stack.
ToolGate is therefore a black-box inference-time controller: its role is not to improve perception directly, but to decide whether paying for an additional perceptual tool output is worthwhile in the current trajectory.

Our main empirical finding is that pre-call gating can substantially reduce tool use and token cost while preserving accuracy.
Across two Qwen3-VL backbones, ToolGate reduces average token cost to 64--69\% of the unrestricted ReAct baseline.
Under cross-domain training, where the gate is trained without trajectories from the five evaluation benchmarks, ToolGate gives small accuracy changes but large efficiency gains; on Qwen3-VL-235B-FP8, it preserves average accuracy while removing most tool calls.
With matched-domain trajectory training on Qwen3-VL-30B, ToolGate additionally improves average accuracy by 1.65 points while reducing token cost by 33\%.
We therefore view ToolGate primarily as an accuracy-preserving cost-control mechanism, with accuracy gains emerging when the baseline policy has sufficient headroom and target-domain trajectories are available.

Our contributions are threefold:
(1) We diagnose a local selectivity problem in ReAct-style VLM agents, showing that useful tool calls are sparse and mixed with many unchanged or harmful calls under the baseline policy.
(2) We introduce \textsc{ToolGate}, a lightweight external pre-call controller that decides whether to execute each proposed perceptual tool call before its output enters the agent context.
(3) We show that ToolGate substantially reduces token cost and executed tool calls while preserving average accuracy in cross-domain settings; with matched-domain trajectory training on Qwen3-VL-30B, it also improves final-answer accuracy.
Code and configurations are available at \url{https://github.com/iamlilAJ/toolgate}.

\section{Related Work}
\label{sec:related}
\paragraph{Tool-augmented language and vision-language agents.}
Recent work augments language models with external tools such as search engines, calculators, code interpreters, visual modules, and APIs~\citep{yao2023react,schick2023toolformer,lu2023chameleon}.
In vision-language settings, systems combine VLMs or language-model planners with OCR, detection, segmentation, cropping, and other visual experts~\citep{yang2023mmreact,wu2023visualchatgpt,gupta2023visualprogramming,suris2023vipergpt}.
Most prior work focuses on letting agents select tools, format calls, follow tool-use protocols, or decide whether a call is appropriate.
Our focus is complementary: we separate \emph{proposing} a tool call from estimating the \emph{utility} of executing it.
ToolGate does not choose which tool to call; given a call already proposed by a vision-language agent, it predicts whether executing it is likely to provide positive marginal value for the current trajectory.

\paragraph{Tool-use failures and inference-time control.}
Recent work has questioned the assumption that tool use is uniformly beneficial.
\citet{zhang2026tools} identify a tool-use tax in LLM agents, decomposing performance loss into prompt-formatting cost, protocol overhead, and tool-execution gain, while \citet{sun2026llm} show that tool necessity can be decoded from pre-generation hidden states and used to steer tool use.
These works are related to broader inference-time control methods that adapt computation, including reducing unnecessary reasoning tokens or shortening chain-of-thought traces when additional deliberation is unlikely to help.
ToolGate studies a complementary granularity: after a vision-language planner has already proposed a concrete perceptual tool call, it decides whether that specific call should be executed before its output enters the context.
Thus, our focus is local pre-call utility estimation rather than task-level tool necessity, termination control, or reasoning-length control~\citep{liu2026perceptual,yu2026causal}.

\paragraph{Selective computation and routing.}
ToolGate is also related to selective computation and model routing, which decide when additional computation is worthwhile~\citep{graves2016adaptive,elbayad2019depth}.
Our setting differs in granularity: the decision is not which model should process an input, but whether a specific proposed tool call inside an agent trajectory should be executed.
Because executing the call changes the evidence available to subsequent reasoning, pre-call gating is closer to selective evidence acquisition than to standard routing or early exiting.

\paragraph{Value of information and active perception.}
The tool-call decision can be seen as a value-of-information problem: an agent should seek external evidence only when the expected gain in final performance exceeds its cost~\citep{howard1966information}.
This links tool use to active perception and experimental design, where observations are chosen adaptively under resource limits~\citep{bajcsy1988active,aloimonos1988active,liu2026perceptual}.
Unlike classical settings, a VLM agent trajectory reveals only one outcome: after a proposed call, we see either the executed or the skipped trajectory, but not both.
Exact counterfactual value of information is therefore unavailable from a single rollout.
We instead train ToolGate with a trajectory-level proxy for decisive calls and evaluate it online in the agent loop.

\section{Local Selectivity Failures in Tool Use}
\label{sec:diagnostic}

Tool-augmented vision–language model (VLM) agents are commonly evaluated solely in terms of final-answer accuracy; however, this aggregate metric obscures the sequence of local tool-use decisions that constitute each reasoning trajectory.  
In this work, we investigate \emph{local tool-call selectivity}: prior to executing a proposed tool invocation, can the agent reliably discriminate between calls that are likely to be beneficial and those that are redundant or potentially detrimental?

Our aim is not to argue that perceptual tools are unnecessary.
Rather, we ask whether the baseline ReAct policy uses them selectively: it should call OCR when text evidence is needed or detection when localization matters, but avoid paying for additional tool outputs when the current trajectory already contains enough evidence.

Unless otherwise specified, all diagnostic evaluations are conducted using the Qwen3-VL-30B ReAct agent and the same tool suite as in our primary experiments.  
Forced-answer probes are employed exclusively for analytical purposes: they capture the agent’s immediate answer preference before and after tool calls, but they are not incorporated into the behavior of the deployed agent.

\subsection{Most proposed calls offer little immediate value}
\label{sec:diagnostic:utility}

We run the baseline ReAct agent on the five evaluation benchmarks and record its forced-answer prediction immediately before and after each executed tool call.
At each point, we apply a \emph{forced-answer probe}, which asks the agent to choose exactly one answer option based only on the trajectory context available so far.
We extract the argmax over the answer options and compare the predictions before and after the tool call.
This classifies each call into four transition types:
\emph{helpful} calls flip the forced-answer argmax from wrong to correct;
\emph{harmful} calls flip it from correct to wrong;
\emph{unchanged-correct} calls leave a correct argmax unchanged;
and \emph{unchanged-wrong} calls leave an incorrect argmax unchanged.

\begin{figure*}[t]
\centering
\includegraphics[width=0.82\linewidth]{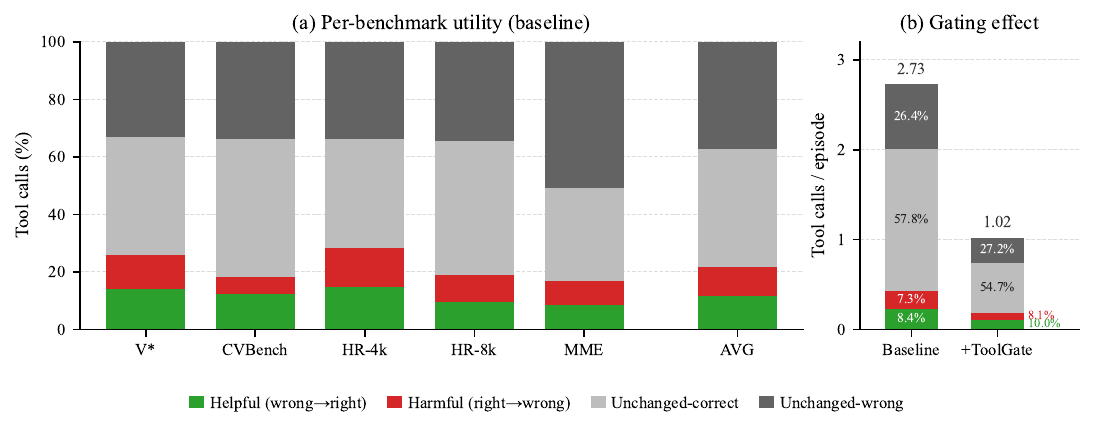}
\caption{
Local tool-call selectivity on Qwen3-VL-30B.
(a) Immediate forced-answer transition distribution for baseline tool calls across five benchmarks.
Helpful calls flip the forced-answer prediction from wrong to correct, while harmful calls flip it from correct to wrong.
Most calls leave the immediate forced-answer prediction unchanged, and helpful and harmful transitions occur at similar rates.
(b) Average executed tool calls per episode before and after ToolGate, decomposed by the same transition types.
ToolGate reduces executed calls from 2.73 to 1.02 per episode, mainly by filtering calls whose immediate forced-answer prediction would remain unchanged.
}
\label{fig:local-selectivity}
\end{figure*}

Figure~\ref{fig:local-selectivity}(a) shows that useful calls are present but sparse.
In aggregate, only 11.8\% of calls convert a wrong forced-answer prediction into a correct one, while 9.9\% convert a correct prediction into a wrong one.
The remaining 78.3\% do not change the forced-answer argmax immediately after the tool output is integrated.
Thus, the baseline policy executes many calls with no immediate measured effect, while positive and negative flips occur at similar rates.
This suggests a local selectivity problem: the issue is not whether tools can help, but whether the agent can identify when a particular proposed call is worth executing.

Figure~\ref{fig:local-selectivity}(b) previews the effect of ToolGate on this decision.
ToolGate reduces executed calls from 2.73 to 1.02 per episode, with the largest absolute reductions coming from unchanged transitions.
This supports the interpretation of ToolGate as a pre-call cost-control mechanism rather than a global tool-removal policy: it prevents many low-value tool outputs from entering the context while retaining a smaller set of executed calls.
We evaluate the resulting accuracy--cost trade-off in Section~\ref{sec:main}.

\subsection{Wrong tool outputs can corrupt correct beliefs}
\label{sec:diagnostic:overtrust}

Why can tool calls hurt?
One possible mechanism is over-trust: once a tool output enters the trajectory, the agent may treat it as stronger evidence than its own visual interpretation.
We examine an automatically annotated diagnostic subset of $1{,}000$ sampled calls, restricted to 802 calls where a VLM annotator judged the tool output verifiable against the image.
When the tool output is wrong and the agent's pre-call forced-answer prediction is already correct, the post-call prediction becomes wrong in 13.9\% of cases ($43/309$).
This suggests that some harmful calls arise because the agent overwrites a correct visual belief with faulty tool output.

This analysis is diagnostic rather than definitive.
Two independent VLM-annotation runs disagree on 35\% of rows on at least one of the \texttt{verifiable} or \texttt{tool\_correctness} fields.
We therefore treat the 13.9\% rate as a rough mechanistic estimate, not as a precise population statistic.

\subsection{Contradictory tool evidence can collapse visual belief}
\label{sec:diagnostic:pendulum}

To test robustness to a clearer contradiction, we run a controlled experiment on Visual Genome~\citep{krishna2017visualgenome}.
For objects certified as present, we compare two programmatic tool stimuli: a consistent stimulus reporting that the object was detected with high confidence, and an adversarial stimulus reporting that no target object was detected.
The stimulus is inserted as text in place of a detector output, isolating the effect of contradictory tool evidence from detector noise.

We evaluate a Naive agent with the standard prompt and a Skeptical agent instructed to distrust tool outputs that contradict visual evidence.
We measure $P_\text{vis}$, the model's probability of answering ``yes'' to whether the target object is visible.

\begin{table}[t]
\centering
\small
\begin{tabular}{l cc}
\toprule
& \textbf{Consistent} & \textbf{Adversarial} \\
\midrule
Naive agent     & 0.993 {\scriptsize [0.992, 0.994]} & \textbf{0.142} {\scriptsize [0.135, 0.148]} \\
Skeptical agent & 0.991 {\scriptsize [0.990, 0.992]} & \textbf{0.461} {\scriptsize [0.450, 0.472]} \\
\bottomrule
\end{tabular}
\caption{Mean $P_\text{vis}$ in a controlled Visual Genome experiment. Under a consistent tool stimulus, both agents retain near-perfect visual belief. Under an adversarial stimulus falsely claiming that a present object is absent, the Naive agent's belief collapses; a skeptical prompt partially recovers belief but does not restore it.}
\label{tab:pendulum}
\end{table}

Table~\ref{tab:pendulum} shows that both agents retain near-perfect visual belief under the consistent stimulus.
Under the adversarial stimulus, the Naive agent's belief collapses to $P_\text{vis}=0.142$.
The Skeptical prompt partially mitigates the collapse, raising $P_\text{vis}$ to 0.461, but does not restore the original belief.
This motivates pre-call control: if a low-value or misleading tool output is likely, preventing it from entering the trajectory may be more reliable than relying on the agent to discount it afterwards.

\subsection{Implication: pre-call local control}
\label{sec:diagnostic:summary}

These analyses point to a local selectivity failure in the baseline ReAct policy.
The agent executes many calls that do not change its immediate answer preference, and some wrong tool outputs can corrupt correct visual beliefs.
This suggests that post-hoc trust is not enough: once misleading tool output enters the context, the agent may fail to discount it.

ToolGate therefore acts before execution.
Instead of asking the agent to recover after seeing a bad or redundant tool result, ToolGate decides whether the result should enter the context at all.
It implements this as a lightweight external execute/skip controller.

\section{ToolGate: Pre-Call Utility Gating}
\label{sec:method}

\begin{figure*}[t]
    \centering
    \includegraphics[width=\linewidth]{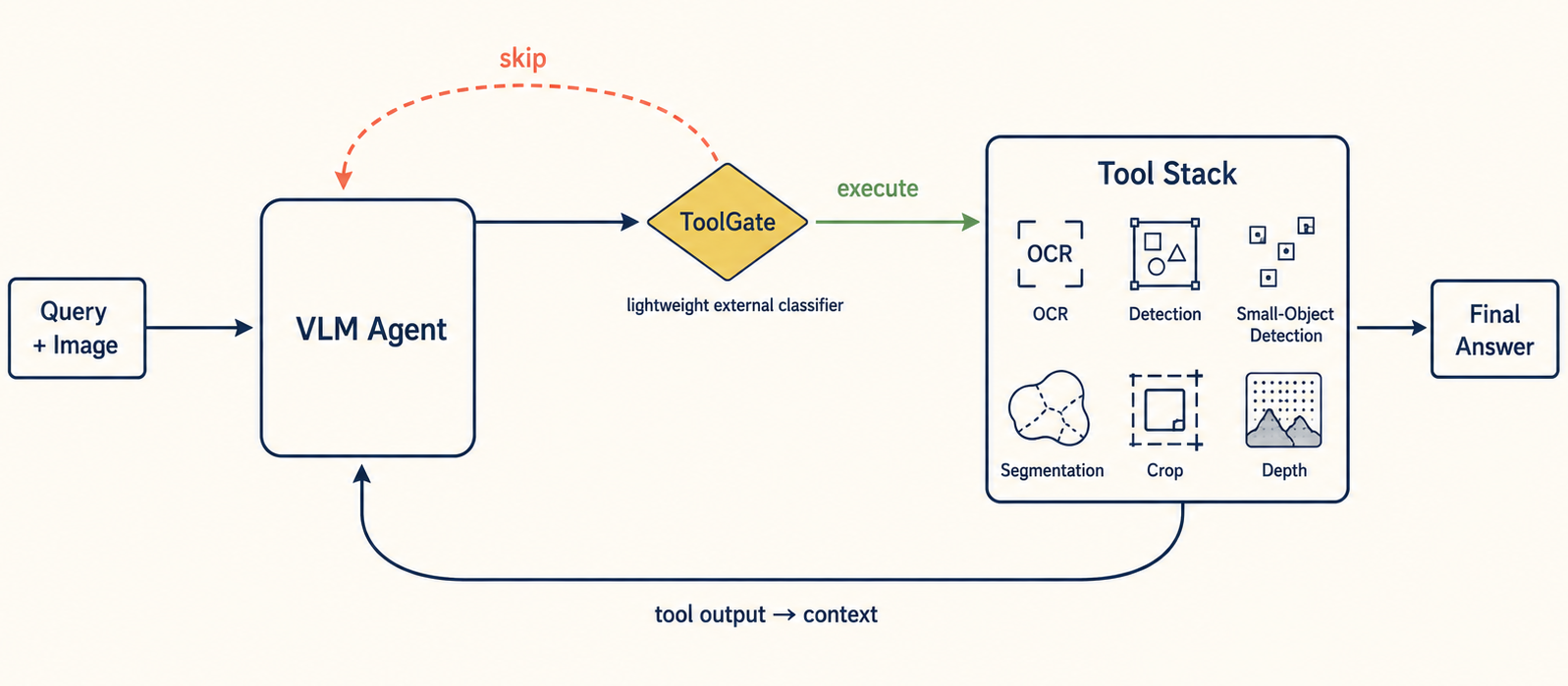}
    \caption{
    ToolGate operates before tool execution.
    The VLM agent proposes a tool call; ToolGate reads the trajectory prefix and pending call, then either executes the call or returns a fixed skip message.
    If executed, the tool output is added back to the agent context; if skipped, no tool output enters the context.
    The gate is external to the VLM and tool stack and does not modify either component.
    }
    \label{fig:toolgate-architecture}
\end{figure*}

We cast tool-call gating as a binary decision made before each proposed tool invocation.
The gate observes the current trajectory state and the pending tool call, then decides whether to execute the call or replace it with a skip message.
Figure~\ref{fig:toolgate-architecture} illustrates the intervention point.

\subsection{Problem formulation}
\label{sec:method:setup}

Let an agent process a query $q$ over turns $t=1,\ldots,T$.
At turn $t$, the agent has produced a trajectory prefix $\pi_t$, consisting of previous reasoning text, executed tool calls, and tool outputs through turn $t-1$.
It then proposes an action $a_t=(\texttt{tool}_t,\texttt{args}_t)$.
A gate $g(\pi_t,a_t)\in\{0,1\}$ decides whether to execute the tool call ($g=1$) or skip it ($g=0$).
If the call is skipped, the environment returns a fixed no-op message and the agent continues.

The ideal decision compares expected final-answer accuracy under two counterfactual continuations:
\begin{equation}
\label{eq:voi}
\begin{aligned}
\mathrm{VoI}(\pi_t,a_t)
&=
\Pr(\hat{y}=y^\star \mid \mathrm{execute}) \\
&\quad -
\Pr(\hat{y}=y^\star \mid \mathrm{skip}) .
\end{aligned}
\end{equation}
where $y^\star$ is the ground-truth answer and $\hat{y}$ is the final answer produced by the completed trajectory.
This quantity is not directly observed in a logged trajectory, because only one continuation is taken.
We therefore treat Eq.~\ref{eq:voi} as the decision-theoretic target and train a practical proxy for whether a call is likely to be decisive for the final answer.

\subsection{Input representation}
\label{sec:method:input}

Figure~\ref{fig:toolgate-architecture} shows where ToolGate operates in the agent loop.
After the VLM agent proposes a tool call, but before the tool is executed, ToolGate receives the current trajectory prefix and the pending call.
The trajectory prefix is a text summary of the interaction so far: the original question, answer choices when present, previous reasoning segments, previous tool calls, and previous tool outputs.
The pending call contains the proposed tool name and its arguments.
ToolGate encodes this text with a frozen sentence-embedding model.

ToolGate is intentionally black-box and lightweight.
It does not see image pixels, hidden states, system prompts, decoding log probabilities, internal uncertainty estimates, or the future tool output.
This design isolates the pre-call decision from the VLM and the tool stack: the gate decides whether the proposed output should enter the context, but it does not change how the agent reasons or how tools execute.

The text embedding is concatenated with a small set of structural features that describe the current tool-use state: the normalized step index, whether the pending call is the first tool call in the trajectory, whether the same tool has already been used, and a one-hot encoding of the proposed tool type.
These features capture simple regularities in the agent loop, such as repeated tool use or tool-specific overuse, without requiring another visual or language reasoning model.

\subsection{Classifier}
\label{sec:method:classifier}

The default gate is an $\ell_2$-regularized logistic regression on the concatenated text embedding and structural features.
Given input $(\pi_t,a_t)$, it outputs a scalar $\hat{p}_t\in[0,1]$, the predicted probability that the call should be executed.
At deployment, the gate executes the call if $\hat{p}_t\geq \tau$ and skips it otherwise.
We use $\tau=0.5$ by default and report threshold sensitivity in Appendix~\ref{sec:appendix:threshold}.

The classifier is intentionally much smaller than the agent.
Its role is not to solve the visual question, but to test whether the local utility of a proposed tool call is predictable from the trajectory state available before execution.

\subsection{Proxy supervision}
\label{sec:method:label}

The ideal label for a proposed tool call is counterfactual: would the final answer be better if the call were executed rather than skipped?
Logged trajectories do not provide this label, because we observe only the executed continuation.
We therefore use forced-answer probes to construct a practical training proxy.

For each logged tool call, we ask the agent for its best answer immediately before and after execution.
Let $c_b,c_a\in\{0,1\}$ denote whether these probed answers are correct, and let $e\in\{0,1\}$ denote whether the final episode answer is correct.
Our default proxy marks a call as \emph{execute-positive} only if it changes the probed answer from wrong to correct ($c_b=0,c_a=1$) in an ultimately successful trajectory ($e=1$).
All other calls are treated as negative for training.

This is a conservative proxy.
It preserves calls that appear clearly decisive, but it may label some delayed-useful calls as negative.
Thus, ToolGate should not be interpreted as estimating true counterfactual value of information.
Instead, we test whether this proxy is sufficient to train an online controller that reduces tool cost while preserving final-answer accuracy.

\section{Experimental Setup}
\label{sec:setup}

We evaluate \textsc{ToolGate} in a ReAct-style vision-language agent with two backbones: Qwen3-VL-30B and FP8-quantized Qwen3-VL-235B.
All deployment comparisons are within-backbone: Baseline and ToolGate share the same serving setup, prompt, decoding configuration, and tool suite.

\paragraph{Agent and tools.}
At each step, the agent outputs either a final answer or a tool call.
The tool suite includes six perceptual tools: OCR, object detection, small-object detection, segmentation, cropping, and depth estimation.
When ToolGate is enabled, it intercepts each tool call and decides whether to execute it or replace it with a fixed skip message before the agent continues.

\paragraph{Benchmarks.}
We conduct evaluations on five held-out vision-language benchmarks:
V\textsuperscript{*}Bench~\citep{vstar},
CV-Bench~\citep{tong2024cambrian1},
HR-Bench-4k and HR-Bench-8k~\citep{hrbench},
and MME-RealWorld-Lite~\citep{zhang2024mmerealworld}, which we subsequently refer to as MME for brevity.
Comprehensive dataset statistics are presented in Appendix~\ref{sec:appendix:data}.

\paragraph{Training data for ToolGate.}
ToolGate is trained offline on trajectories produced by the unrestricted ReAct agent.
Each proposed tool call in a logged trajectory becomes one training instance: the input is the serialized trajectory prefix and pending call, and the label is derived from forced-answer probes before and after tool execution together with the final episode outcome (Section~\ref{sec:method:label}).
The forced-answer probes are used only for offline label construction and diagnostics, not during deployment.

We consider two training regimes.
In the \emph{in-domain} regime, the gate is trained on info-gain trajectories from the same five benchmark families used for evaluation.
This tests whether ToolGate can learn an execute/skip controller when target-task trajectories are available; it should be viewed as a matched-domain operating point, not a zero-shot generalization result.
In the \emph{cross-domain} regime, the gate is trained only on out-of-domain VQA sources, using no trajectories from the five evaluation benchmarks.
For 30B, the cross-domain gate is trained on GQA-derived info-gain trajectories~\citep{hudson2019gqa}.
For 235B-FP8, it is trained on six out-of-domain VQA sources: A-OKVQA~\citep{schwenk2022aokvqa}, GQA~\citep{hudson2019gqa}, TextVQA~\citep{singh2019textvqa}, DocVQA~\citep{mathew2021docvqa}, SEED-Bench~\citep{li2024seedbench}, and RSVL-MQA~\citep{zi2025rsvlmqa}.

\paragraph{Metrics.}
We report final-answer accuracy, token cost relative to the unrestricted ReAct baseline, and tool calls per episode.
Token cost includes agent interaction tokens and tool-output tokens within the ReAct loop.
ToolGate is deployed online: each gate decision alters the subsequent trajectory rather than replaying a fixed log.
All agent generations use deterministic decoding with temperature 0, and the default threshold is $\tau=0.5$.

\section{Results}
\label{sec:main}

\subsection{Main accuracy--cost trade-off}
\label{sec:main:scale}

\begin{table*}[t]
\centering
\small
\setlength{\tabcolsep}{4pt}
\begin{tabular}{l l l rrrrr r rr}
\toprule
\textbf{Backbone} & \textbf{Training} & \textbf{Method}
& \textbf{V\textsuperscript{*}} & \textbf{CV} & \textbf{HR4k} & \textbf{HR8k} & \textbf{MME}
& \textbf{Acc} & \textbf{Cost} & \textbf{Calls} \\
\midrule
30B & --- & Baseline
& 75.9 & 78.8 & 76.1 & 72.6 & 50.8
& 70.83 & 100\% & 2.73 \\

30B & Cross-domain & + ToolGate
& 76.4 & 77.6 & 76.3 & 74.3 & 51.1
& 71.15 & 64\% & 1.00 \\

30B & In-domain & + ToolGate
& \textbf{79.1} & \textbf{80.0} & \textbf{77.3} & \textbf{75.5} & 50.4
& \textbf{72.48} & 67\% & 1.02 \\

\addlinespace
235B-FP8 & --- & Baseline
& 80.6 & 82.7 & 81.9 & 78.2 & 52.6
& 75.22 & 100\% & 1.34 \\

235B-FP8 & Cross-domain & + ToolGate
& 80.1 & \textbf{83.7} & 80.5 & 77.0 & \textbf{55.1}
& \textbf{75.30} & 69\% & 0.20 \\
\bottomrule
\end{tabular}
\caption{Main online deployment results. Accuracy is reported per benchmark and as an unweighted average over five benchmarks. Cost is average token cost relative to the unrestricted ReAct baseline for the same backbone; Calls is executed tool calls per episode. Cross-domain gates are trained without trajectories from the five evaluation benchmarks. The 30B in-domain gate is trained on trajectories from the same benchmark families and should be interpreted as a matched-domain operating point rather than zero-shot transfer.}
\label{tab:main}
\end{table*}

Table~\ref{tab:main} summarizes the main online deployment results.
The most consistent effect of ToolGate is efficiency: across the reported settings, it reduces average token cost to 64--69\% of the unrestricted ReAct baseline.
These savings come from fewer executed perceptual tool calls rather than shorter final answers alone.

Under cross-domain training, ToolGate primarily acts as an accuracy-preserving cost controller.
On Qwen3-VL-30B, a gate trained only on GQA-derived trajectories reduces average token cost to 64\% of the baseline and reduces executed calls from 2.73 to roughly 1.00 per episode, while average accuracy changes from 70.83 to 71.15.
On Qwen3-VL-235B-FP8, a gate trained on six out-of-domain VQA sources reduces token cost to 69\% and executed calls from 1.34 to 0.20 per episode, while average accuracy remains essentially unchanged.
Thus, the strict cross-domain results should be read mainly as evidence for cost reduction under preserved accuracy, not as large accuracy gains.

Matched-domain training can provide an additional accuracy upside.
On Qwen3-VL-30B, the in-domain gate raises average accuracy from 70.83 to 72.48 while reducing token cost to 67\% of the baseline and executed calls from 2.73 to 1.02 per episode.
Because this gate is trained on trajectories from the same benchmark families used for evaluation, we view it as a matched-domain operating point that illustrates the potential of pre-call gating when target-task trajectories are available.
We do not treat this row as evidence of zero-shot generalization.

\subsection{Prompting the agent to self-regulate is insufficient}
\label{sec:main:alternatives}

\begin{table}[t]
\centering
\small
\begin{tabular}{l rrr}
\toprule
\textbf{Method (30B)} & \textbf{AVG Acc} & \textbf{Tokens} & \textbf{Calls} \\
\midrule
Baseline               & 70.8 & 100\% & 2.73 \\
VoI self-report         & 60.0 & 126\% & 5.88 \\
Sparse prompt           & 69.8 & 99\%  & 2.40 \\
ToolGate                & \textbf{72.5} & 67\% & 1.02 \\
\bottomrule
\end{tabular}
\caption{Agent-side prompting vs. external gating on Qwen3-VL-30B. ``Calls'' is the number of executed tool calls per episode. ToolGate uses the in-domain gate from Table~\ref{tab:main}.}
\label{tab:alternatives}
\end{table}

Table~\ref{tab:alternatives} compares ToolGate with two prompt-based controls.
Asking the agent to self-report whether a tool is worthwhile increases tool use and sharply reduces accuracy.
A sparse-use instruction slightly reduces calls but leaves token cost almost unchanged and does not improve accuracy.
This suggests that the cost savings in Table~\ref{tab:main} do not arise simply from telling the model to use fewer tools; an external controller provides a more reliable execution boundary.

\section{Ablations and Analysis}
\label{sec:ablation}

\subsection{Proxy label formulation}
\label{sec:ablation:label}

The ideal value of a tool call is counterfactual: it depends on how the trajectory would unfold if the call were executed versus skipped.
Since logged trajectories contain only the observed continuation, ToolGate is trained with proxy labels derived from forced-answer probes and final episode correctness.
We compare two formulations.
The \emph{tool-useful} label uses immediate before/after changes in forced-answer correctness and confidence.
The \emph{execute-positive} proxy additionally conditions on final episode correctness and marks a call as execute-positive only when it is a decisive wrong-to-right flip in a successful trajectory.
We use execute-positive as the default label because the gating decision is ultimately about preserving calls that appear necessary for final correctness, not merely calls that change local confidence.

This label is intentionally conservative and should not be interpreted as true causal value of information.
A negative label does not prove that a call has zero downstream value; it only means that the call did not match our execute-positive proxy in the logged trajectory.
We use the proxy as a practical supervision signal for cost control, and evaluate its effect through online deployment rather than only offline classification.

\subsection{How much signal comes from tool identity?}
\label{sec:ablation:features}

\begin{table}[t]
\centering
\small
\setlength{\tabcolsep}{4pt}
\begin{tabular}{lrrrr}
\toprule
\textbf{Variant} & \textbf{Acc} & $\Delta$ & \textbf{Cost} & \textbf{Calls} \\
\midrule
Baseline        & 70.83 & ---   & 100\% & 2.73 \\
Full            & 72.48 & +1.65 & 67\%  & 1.02 \\
Text-only       & 70.69 & $-0.14$ & 75\% & 0.98 \\
Struct-only     & 71.81 & +0.98 & 69\%  & 0.97 \\
Tool-only       & 72.14 & +1.31 & 75\%  & 1.34 \\
\bottomrule
\end{tabular}
\caption{Feature ablation on Qwen3-VL-30B using the in-domain gate. Tool identity is already a strong signal, indicating systematic tool-level inefficiencies in the baseline policy. The full gate gives the best overall accuracy--cost trade-off. Cost is token cost relative to the unrestricted ReAct baseline; Calls is executed tool calls per episode.}
\label{tab:feature-ablation}
\end{table}

Table~\ref{tab:feature-ablation} shows that tool identity alone is a strong baseline.
This is an important finding: a substantial part of the baseline inefficiency is systematic at the tool level, rather than requiring deep semantic understanding of the trajectory.
However, the full gate still improves the deployed trade-off over the tool-only variant, achieving higher accuracy with lower token cost and fewer executed calls.
Thus, trajectory state provides additional signal beyond a static per-tool prior, although the gain over tool-only is modest.

We also test whether ToolGate simply thresholds the agent's current confidence.
Across $15{,}782$ tool calls, the correlation between the gate score and the agent's pre-call probability assigned to the ground-truth answer is weak: Pearson $r=-0.19$ and Spearman $\rho=-0.22$.
This suggests that pre-call confidence alone does not explain the gate's decisions.

\subsection{Negative findings}
\label{sec:ablation:negatives}

We explored richer input signals, including image-derived features, vision-language encoder features, and model log-probability features.
These variants did not improve the deployed accuracy--cost trade-off over the lightweight text-and-structure gate.
This supports the use of a simple external controller for our setting: useful cost-control signals can be extracted from trajectory text, tool identity, and structural features, while heavier features do not automatically improve online behavior.
Threshold sweeps are reported in Appendix~\ref{sec:appendix:threshold}; increasing the threshold monotonically reduces tool calls, while moderate thresholds give the best 30B accuracy.

\section{Discussion}
\label{sec:discussion}

\paragraph{Accuracy-preserving cost control.}
The most robust effect of ToolGate is not a large accuracy improvement, but a substantial reduction in tool use and token cost under comparable accuracy.
This matters for tool-augmented VLM agents because each perceptual tool call adds latency, tool-output tokens, and additional context for subsequent reasoning.
ToolGate acts as a pre-call controller that prevents many low-value tool outputs from entering the context.
In the 30B matched-domain setting, this cost reduction is accompanied by an accuracy gain; in stricter cross-domain settings, the main benefit is efficiency under preserved accuracy.

\paragraph{Local selectivity, not tool removal.}
Our results do not imply that perceptual tools are useless.
Figure~\ref{fig:local-selectivity} shows that helpful calls exist, but they are sparse and mixed with many unchanged or harmful calls under the baseline ReAct policy.
ToolGate therefore targets a local execute/skip decision rather than global tool removal: it reduces executed calls while preserving a smaller set of calls that the gate predicts to be worth paying for.

\paragraph{Training regime and backbone regime.}
The largest 30B gain comes from an in-domain gate trained on trajectories from the same benchmark families used for evaluation, so it should be viewed as a matched-domain operating point rather than zero-shot transfer.
Cross-domain results are stricter and should be interpreted mainly as cost-saving results with approximately preserved accuracy.
An additional 235B-FP8 in-domain gate reduces token cost but hurts accuracy (Appendix~\ref{sec:appendix:235b-indomain}), indicating that matched-domain data is not automatically better when the execute-positive class is sparse and difficult to infer from text-only trajectory features.

\paragraph{Proxy supervision.}
ToolGate is not a causal estimator of true counterfactual value of information.
Its labels are proxies derived from forced-answer probes and final episode correctness, so positive labels should be read as execute-positive supervision rather than proof that all negative calls have zero value.
The contribution is therefore not exact VoI estimation, but evidence that a lightweight external controller can reduce tool calls and token cost while preserving online task performance.
The supervision also requires offline forced-answer probes, so label collection is a one-time data cost for a given agent/tool setup.

\section{Conclusion}
\label{sec:conclusion}

We studied pre-call control for tool-augmented vision-language agents.
Across five benchmarks, a ReAct-style VLM agent proposes many perceptual tool calls whose immediate forced-answer effect is unchanged, while helpful and harmful transitions occur at similar rates.
We introduced ToolGate, a lightweight external controller that decides whether to execute or skip each proposed tool call before its output enters the agent context.
Across two Qwen3-VL backbones, ToolGate substantially reduces token cost and executed tool calls while preserving average accuracy; in a matched-domain 30B setting, it also improves final-answer accuracy.
These results suggest that tool-augmented VLM agents need not only access to perceptual tools, but also explicit inference-time control over when tool outputs are worth paying for.

\section*{Limitations}

This study is limited to a ReAct-style VLM agent with one fixed perceptual tool suite and two backbones from the same model family.
Although the results suggest that pre-call selectivity is useful in this setting, they do not establish that the same gate will transfer unchanged to other agent planners, tool APIs, modalities, or domains.
Future work should evaluate ToolGate-style controllers on different VLM families and agent protocols.

ToolGate is trained with a proxy for counterfactual value of information rather than true paired execute/skip labels.
The proxy is derived from forced-answer probes and final episode correctness, and it may miss calls whose value is delayed, distributed across multiple future steps, or reflected in confidence rather than an immediate argmax flip.
Thus, our results should be interpreted as evidence that useful pre-call selectivity can be learned, not as evidence that the classifier estimates exact causal VoI.

The diagnostic annotation of tool correctness is automatic and noisy.
We use a VLM judge to annotate whether tool outputs are verifiable and correct, and two annotation runs disagree on a non-trivial fraction of examples.
We therefore treat the over-trust analysis as mechanistic evidence rather than a precise measurement of the frequency of harmful tool outputs.

Finally, a pre-call gate can make harmful mistakes.
If it skips too aggressively, it may suppress evidence required for OCR-heavy, small-object, or visually cluttered examples.
If it executes too aggressively, misleading tool outputs may still enter the context.
For high-stakes applications, ToolGate should not be used as a substitute for human oversight or domain-specific validation.

\bibliography{custom}
\appendix

\section{Implementation Details}
\label{sec:appendix:impl}

\paragraph{Tool specifications.}
We use a fixed perceptual tool suite throughout the experiments.
For open-vocabulary object detection, we use Grounding DINO~\citep{liu2024groundingdinomarryingdino}; it takes an image and a text prompt and returns bounding boxes, which are drawn on the image and returned to the agent together with coordinates.
For segmentation, we use SAM 2~\citep{ravi2024sam2segmentimages}; masks are overlaid on the image and returned to the agent.
For monocular depth estimation, we use Depth Anything~\citep{yang2024depthanythingunleashingpower}; the agent receives a relative depth heatmap together with the original image.
For OCR, we use MinerU~\citep{niu2025mineru25decoupledvisionlanguagemodel}, which processes an image region and returns structured text.
The remaining exposed tools are cropping and small-object detection variants built on the same image-processing pipeline.

\paragraph{Trajectory prefix format.}
For a proposed call at step $t$, ToolGate serializes the current trajectory into text.
The prefix concatenates the query and answer choices under \texttt{[Q]}, each completed thinking segment \texttt{[T$_i$]} truncated to 200 characters, each previous tool call \texttt{[TOOL$_i$]} rendered as \texttt{name(args) -> output} with arguments truncated to 80 characters and outputs to 150 characters, and the pending call \texttt{[PENDING]} rendered as \texttt{name(args)}.
The full prefix is left-truncated to 1{,}500 characters, while \texttt{[Q]} and \texttt{[PENDING]} are always retained.
These bounds were set to fit the MiniLM-L6-v2 context window with margin and were not tuned.

\paragraph{Structural features.}
The structural feature vector has nine dimensions:
$s_1=t/10$ is the normalized step index;
$s_2\in\{0,1\}$ indicates whether the pending call is the first tool call in the trajectory;
$s_3\in\{0,1\}$ indicates whether the same tool has already appeared in the trajectory;
and $s_{4:9}\in\{0,1\}^6$ is a one-hot encoding over the six tools exposed to the planner:
\texttt{ocr}, \texttt{detection}, \texttt{detection\_small\_object}, \texttt{segmentation}, \texttt{cropping}, and \texttt{depth\_estimation}.

\paragraph{Encoder and classifier.}
The default ToolGate uses \texttt{sentence-transformers/all-MiniLM-L6-v2} as a frozen sentence encoder ($d=384$, mean pooling).
The embedding is concatenated with the nine structural features and passed to a logistic-regression classifier.
We train with $\ell_2$ regularization, $C=1.0$, \texttt{class\_weight=balanced}, the LBFGS solver, and a 1000-iteration cap.
The default deployment threshold is $\tau=0.5$.

\paragraph{Forced-answer probing.}
To label logged trajectories, we probe the agent immediately before and after each executed tool call.
At each probe point, we append the following user turn:
\begin{quote}\itshape
Based on all information available to you so far, what is your best answer to the original question? You MUST choose exactly one option letter. Respond with ONLY the letter (A, B, C, or D), nothing else.
\end{quote}
We call the model with \texttt{max\_tokens=1}, \texttt{logprobs=True}, and \texttt{top\_logprobs=20}, then extract probabilities for the option-letter tokens.
These probes are used only for offline label construction and diagnostics; they are not part of the deployed agent.

\paragraph{Proxy-label derivation.}
Let $c_b,c_a\in\{0,1\}$ denote whether the forced-answer argmax is correct before and after the call, and let $e\in\{0,1\}$ denote whether the episode's final answer is correct.
We compare two proxy labels.
The \emph{tool-useful} label is positive if the call changes $c_b=0$ to $c_a=1$, or if $c_b=c_a=1$ and the probability assigned to the ground-truth answer increases by more than $\delta=0.1$.
The \emph{execute-positive} proxy marks a call as execute-positive only when $c_b=0$, $c_a=1$, and the episode is ultimately successful.
We use execute-positive as the default label.
This is a practical proxy for training a cost-control gate, not a measurement of true counterfactual value of information.

\paragraph{Inference-time cost.}
A single ToolGate query consists of one sentence-embedding forward pass over the serialized prefix and one logistic-regression dot product.
In our deployment this runs in well under 10 ms on a single CPU core, making the gate substantially cheaper than a perceptual tool invocation.

\paragraph{Automatically annotated diagnostic subset.}
The 1{,}000 tool calls in Section~\ref{sec:diagnostic:overtrust} were sampled from the info-gain-labeled trajectory pool.
Each call was annotated automatically by a VLM judge, which received the original image, question, answer choices, tool name, tool arguments, and tool output.
The judge labeled whether the tool output was verifiable and, conditional on verifiability, whether it was factually correct.
To assess annotation stability, we ran the VLM annotator twice independently.
The two runs disagree on 35\% of rows on at least one of the \texttt{verifiable} or \texttt{tool\_correctness} fields.
We therefore treat rates derived from this subset as diagnostic estimates rather than precise population statistics.
The before/after answer correctness values are not annotation-derived; they come from forced-answer probing.

\paragraph{Pendulum experiment.}
The Visual Genome probe set was constructed by filtering object annotations for COCO-mapped categories and retaining examples where a target object is present.
Each example is evaluated under two programmatic tool messages.
The consistent message reports that the target object is detected with high confidence; the adversarial message reports that no target object is detected.
$P_\text{vis}$ is computed from the first-token log probability of \texttt{yes} under a binary visibility question.
Confidence intervals use a normal approximation.

\section{Dataset and Benchmark Statistics}
\label{sec:appendix:data}

Table~\ref{tab:eval-bench-stats} summarizes the five evaluation benchmarks.
All reported accuracies are computed on the full benchmark subsets listed below.
Averages are unweighted means over the five benchmark-level accuracies.

\begin{table*}[t]
\centering
\small
\setlength{\tabcolsep}{4pt}
\begin{tabular}{l l r l}
\toprule
\textbf{Benchmark} & \textbf{Task focus} & \textbf{Examples} & \textbf{Resolution} \\
\midrule
V\textsuperscript{*}Bench & Fine-grained visual search & 191 & $\leq$2K \\
CV-Bench & Count/relation/depth/distance & 2{,}638 & 256px \\
HR-Bench-4k & High-resolution perception & 800 & 4K \\
HR-Bench-8k & High-resolution perception & 800 & 8K \\
MME-RealWorld-Lite & Real-world perception/reasoning & 1{,}919 & $\leq$7K \\
\midrule
Total & --- & 6{,}348 & --- \\
\bottomrule
\end{tabular}
\caption{Evaluation benchmark statistics.}
\label{tab:eval-bench-stats}
\end{table*}

Table~\ref{tab:probe-training-stats} summarizes the offline trajectory pools used to train ToolGate.
The VLM agent is frozen throughout; only the external probe is trained.
Each proposed tool call in a logged trajectory becomes one training instance.
The positive class corresponds to execute-positive calls under the execute-positive proxy label.
All probes use a frozen MiniLM-L6-v2 sentence encoder, nine structural features, and class-balanced logistic regression.

\begin{table*}[t]
\centering
\small
\setlength{\tabcolsep}{3.5pt}
\begin{tabular}{l l r r r}
\toprule
\textbf{Probe} & \textbf{Training data} & \textbf{Calls} & \textbf{Pos. rate} & \textbf{CV AUC} \\
\midrule
30B in-domain
& 5 eval benches (4{,}207 ep.)
& 11{,}596 & 5.5\% & 0.674 \\

30B cross-domain
& GQA only (OOD)
& 11{,}298 & 23.4\% & 0.669 \\

235B cross-domain
& 6 OOD VQA sources
& 3{,}571 & 4.2\% & 0.614 \\
\bottomrule
\end{tabular}
\caption{ToolGate training-pool statistics. ``Calls'' denotes labeled proposed tool calls. ``Pos. rate'' is the positive-label rate under the execute-positive proxy: a call is positive when it produces an immediate wrong-to-right forced-answer flip in an ultimately successful trajectory. ``CV AUC'' is 5-fold cross-validation AUROC on the corresponding training pool.}
\label{tab:probe-training-stats}
\end{table*}

The 30B in-domain pool contains trajectories from the same benchmark families used for evaluation:
V\textsuperscript{*}Bench (191 episodes), CV-Bench (500), HR-Bench-4k (797), HR-Bench-8k (800), and MME-RealWorld-Lite (1{,}919).

The 30B cross-domain pool is trained only on GQA-derived trajectories and contains no examples from the five evaluation benchmarks.
The 235B cross-domain pool contains trajectories from six out-of-domain VQA sources:
A-OKVQA (2{,}000 episodes), GQA (1{,}140), TextVQA (636), DocVQA (500), SEED-Bench (385), and RSVLM-QA (355).
No evaluation-benchmark trajectories are used to train the 235B cross-domain gate reported in the main table.

\section{Additional 235B-FP8 In-Domain Result}
\label{sec:appendix:235b-indomain}

We additionally trained a 235B-FP8 in-domain ToolGate on trajectories collected from the same five benchmark families used for evaluation.
The probe uses the same MiniLM-L6-v2 encoder, nine structural features, class-balanced logistic regression, execute-positive proxy label, and threshold $\tau=0.5$ as the main gates.
The training pool contains 5{,}201 episodes and yields 5-fold CV AUROC of 0.61.

\begin{table*}[t]
\centering
\small
\setlength{\tabcolsep}{4pt}
\begin{tabular}{lrrrrrrr}
\toprule
\textbf{Setting} & \textbf{V\textsuperscript{*}} & \textbf{CV} & \textbf{HR4k} & \textbf{HR8k} & \textbf{MME} & \textbf{AVG} & \textbf{Cost} \\
\midrule
Baseline & 80.6 & 82.7 & 81.9 & 78.2 & 52.6 & 75.22 & 100\% \\
Cross-domain & 80.1 & 83.7 & 80.5 & 77.0 & 55.1 & 75.30 & 69\% \\
In-domain & 75.9 & 83.8 & 79.1 & 76.4 & 54.3 & 73.90 & 63\% \\
\bottomrule
\end{tabular}
\caption{Additional 235B-FP8 in-domain result. The in-domain gate reduces token cost but underperforms both the unrestricted baseline and the cross-domain gate in accuracy. Cost is average token cost relative to the baseline.}
\label{tab:235b-indomain}
\end{table*}

The 235B-FP8 in-domain gate reduces token cost to 63\% of the baseline but lowers average accuracy by 1.32 points.
This contrasts with the 30B setting, where in-domain training gives the strongest result.
We interpret this as evidence that matched-domain trajectory data is not automatically beneficial.
For the stronger backbone, the execute-positive class is sparse and the text-only probe has limited access to the visual and tool-output information needed to identify the few calls that remain decisive.
As a result, the in-domain gate can overfit noisy benchmark-specific trajectory cues and over-skip useful evidence.
The cross-domain 235B-FP8 gate is therefore the main 235B operating point reported in Table~\ref{tab:main}.

\section{Sparse-Prompt Per-Benchmark Results}
\label{sec:appendix:sparse}

The single-line instruction appended to the planner prompt was:
\emph{``Use tools sparingly --- only call a tool when you genuinely cannot answer from the image alone.''}

\begin{table}[t]
\centering
\small
\begin{tabular}{l rr r r}
\toprule
\textbf{Benchmark} & \textbf{Baseline} & \textbf{Sparse} & $\Delta$ Acc & \textbf{Cost} \\
\midrule
V\textsuperscript{*}Bench & 75.9 & 70.2 & $-5.8$ & 109\% \\
CV-Bench & 78.8 & 78.0 & $-0.8$ & 102\% \\
HR-Bench-4k & 76.1 & 78.6 & $+2.5$ & 95\%  \\
HR-Bench-8k & 72.6 & 75.3 & $+2.8$ & 99\%  \\
MME-RealWorld-Lite & 50.8 & 47.1 & $-3.7$ & 93\%  \\
\midrule
AVG & 70.8 & 69.8 & $-1.0$ & 99\%  \\
\bottomrule
\end{tabular}
\caption{Sparse-prompt per-benchmark results on Qwen3-VL-30B. Cost is token cost relative to the unrestricted ReAct baseline.}
\label{tab:sparse-perbench}
\end{table}

\section{Feature Ablation Details}
\label{sec:appendix:feature-ablation}

\begin{table}[t]
\centering
\small
\begin{tabular}{l rr rr r}
\toprule
\textbf{Variant} & \textbf{AUROC} & \textbf{AVG Acc} & $\Delta$ & \textbf{Cost} & \textbf{Calls} \\
\midrule
Baseline           & ---   & 70.83 & ---   & 100\% & 2.73 \\
Full               & 0.663 & 72.48 & +1.65 & 67\%  & 1.02 \\
Text-only          & 0.643 & 70.69 & $-0.14$ & 75\% & 0.98 \\
Struct-only        & 0.620 & 71.81 & +0.98 & 69\%  & 0.97 \\
Tool-only          & 0.568 & 72.14 & +1.31 & 75\%  & 1.34 \\
\bottomrule
\end{tabular}
\caption{Feature ablation on Qwen3-VL-30B, including offline AUROC. Cost is token cost relative to the unrestricted ReAct baseline; Calls is executed tool calls per episode.}
\label{tab:feature-ablation-appendix}
\end{table}

\section{Threshold Sensitivity}
\label{sec:appendix:threshold}

ToolGate uses a deployment-time threshold $\tau$: a proposed tool call is executed when the predicted execution probability is at least $\tau$ and skipped otherwise.
Table~\ref{tab:threshold-appendix} reports threshold sweeps on both backbones.
Increasing $\tau$ monotonically reduces the number of executed tool calls.
On Qwen3-VL-30B, moderate thresholds give the best accuracy, with $\tau=0.4$ and $\tau=0.5$ performing similarly.
On Qwen3-VL-235B-FP8, accuracy remains within a narrow range while tool calls decrease sharply.

\begin{table}[t]
\centering
\small
\setlength{\tabcolsep}{4pt}
\begin{tabular}{lrrrr}
\toprule
\textbf{Backbone} & $\boldsymbol{\tau}$ & \textbf{Acc} & \textbf{Cost} & \textbf{Calls} \\
\midrule
30B & baseline & 70.83 & 100\% & 2.73 \\
30B & 0.3 & 71.44 & 80\% & 1.82 \\
30B & 0.4 & \textbf{72.49} & 74\% & 1.45 \\
30B & 0.5 & 72.48 & 67\% & 1.02 \\
30B & 0.6 & 71.67 & 61\% & 0.67 \\
30B & 0.7 & 71.47 & 54\% & 0.31 \\
\midrule
235B-FP8 & baseline & 75.22 & 100\% & 1.34 \\
235B-FP8 & 0.3 & 74.46 & 80\% & 0.51 \\
235B-FP8 & 0.4 & 74.44 & 74\% & 0.33 \\
235B-FP8 & 0.5 & \textbf{75.30} & 69\% & 0.20 \\
235B-FP8 & 0.6 & 75.02 & 66\% & 0.11 \\
235B-FP8 & 0.7 & 74.37 & 64\% & 0.05 \\
\bottomrule
\end{tabular}
\caption{Threshold sensitivity. Cost is token cost relative to the unrestricted ReAct baseline; Calls is executed tool calls per episode.}
\label{tab:threshold-appendix}
\end{table}
\section{Artifacts, Licenses, and Compute}
\label{sec:appendix:artifacts}

\paragraph{Scientific artifacts.}
Our experiments use existing models, tools, and benchmarks released by their respective creators.
The VLM backbones are Qwen3-VL-30B and Qwen3-VL-235B-FP8~\citep{bai2025qwen3vl}.
The perceptual tool suite uses Grounding DINO~\citep{liu2024groundingdinomarryingdino}, SAM 2~\citep{ravi2024sam2segmentimages}, Depth Anything~\citep{yang2024depthanythingunleashingpower}, and MinerU~\citep{niu2025mineru25decoupledvisionlanguagemodel}.
The evaluation benchmarks are V\textsuperscript{*}Bench~\citep{vstar}, CV-Bench~\citep{tong2024cambrian1}, HR-Bench~\citep{hrbench}, and MME-RealWorld-Lite~\citep{zhang2024mmerealworld}.
We also use Visual Genome~\citep{krishna2017visualgenome} for the controlled contradictory-tool experiment.

\paragraph{Licenses and intended use.}
We use all artifacts for research evaluation of tool-augmented VLM agents.
For models, tools, and datasets obtained from public repositories or benchmark releases, we follow the licenses, terms of use, and redistribution conditions specified by the original artifact providers.
We do not redistribute the original model weights or benchmark data as part of this submission.
Our released artifacts, if any, will contain only code, configuration files, aggregate statistics, and derived trajectory metadata permitted by the corresponding artifact terms.

\paragraph{Data privacy and content.}
We do not collect new human-subject data.
The evaluation uses existing public vision-language benchmarks and public image datasets.
Some public image benchmarks may contain people or scene content; we use them only for aggregate research evaluation and do not attempt to identify individuals.
The ToolGate training data consists of derived agent trajectories, tool-call records, forced-answer probes, and aggregate labels, rather than newly collected personal annotations.

\paragraph{Compute.}
All VLM weights are frozen during our experiments; only the lightweight ToolGate classifier is trained.
Training ToolGate requires sentence-embedding extraction and logistic-regression fitting over logged tool-call instances, which is inexpensive relative to VLM inference.
The main computational cost comes from collecting ReAct trajectories, executing perceptual tools, and running offline forced-answer probes for label construction.
For Qwen3-VL-30B, we evaluate the baseline and gated agents on the full five-benchmark suite.
For Qwen3-VL-235B-FP8, we use the same benchmark suite and an FP8-quantized serving setup for all within-backbone comparisons.
Exact total GPU hours were not logged uniformly across all trajectory-collection jobs; therefore, we report token cost and executed tool calls per episode as the primary compute-efficiency metrics in the main paper.

\section{Qualitative Examples}
\label{sec:appendix:qualitative}

\begin{figure*}[t]
\centering

\begin{subfigure}{0.95\linewidth}
    \centering
    \includegraphics[width=0.9\linewidth]{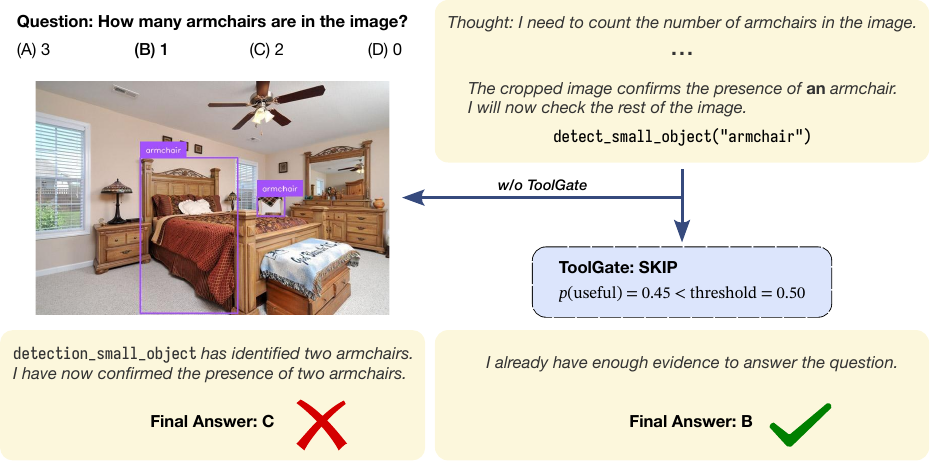}
    \caption{}
\end{subfigure}

\vspace{0.8em}

\begin{subfigure}{0.95\linewidth}
    \centering
    \includegraphics[width=0.9\linewidth]{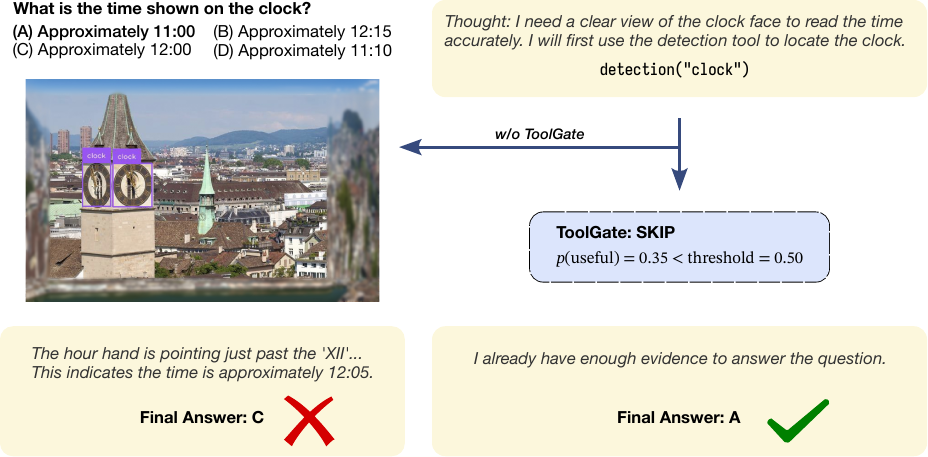}
    \caption{}
\end{subfigure}

\vspace{0.8em}

\begin{subfigure}{0.95\linewidth}
    \centering
    \includegraphics[width=0.9\linewidth]{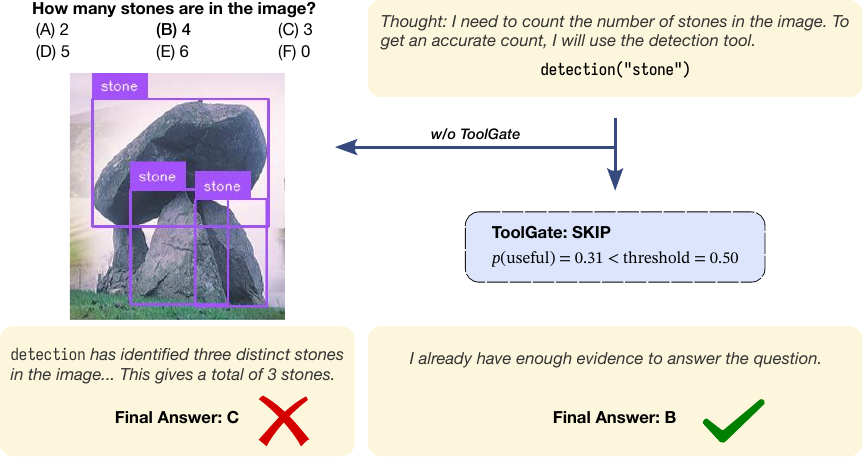}
    \caption{}
\end{subfigure}

\label{fig:qualitative-examples}
\end{figure*}

\end{document}